\documentclass[letterpaper]{article} 
\usepackage{aaai25}  
\usepackage{times}  
\usepackage{helvet}  
\usepackage{courier}  
\usepackage[hyphens]{url}  
\usepackage{graphicx} 
\usepackage{natbib}  
\usepackage{caption} 
\newcommand{\method}[1]{{\textsc{GMT}}} 
\newcommand{\methodfull}[1]{{Gradient-Mask Tuning}} 
\title{\methodfull{} Elevates the Upper Limits of LLM Performance}

\title{Enhancing Large Language Model Performance with \\ Gradient-Based Parameter Selection}
\author{
    Haoling Li\textsuperscript{\rm 1}\thanks{~~Work done during his internship at Microsoft.}, Xin Zhang\textsuperscript{\rm 2}\thanks{~~Corresponding author.}, Xiao Liu\textsuperscript{\rm 2}, Yeyun Gong\textsuperscript{\rm 2}, Yifan Wang\textsuperscript{\rm 1}, Qi Chen\textsuperscript{\rm 2}, Peng Cheng\textsuperscript{\rm 2}
}
\affiliations{
    \textsuperscript{\rm 1}Tsinghua University \\
    \textsuperscript{\rm 2} Microsoft Research \\
    li-hl23@mails.tsinghua.edu.cn, xinzhang3@microsoft.com
}

\usepackage{algorithm}
\usepackage{algorithmic}

\usepackage{enumitem}  
\usepackage{subcaption}
\usepackage{multirow}
\usepackage{tabularx}
\usepackage{booktabs}
\usepackage{colortbl}
\usepackage{threeparttable}
\usepackage{listings}
\usepackage{amsmath}
\usepackage{amssymb} 
\usepackage[noabbrev,capitalize]{cleveref}

\crefname{equation}{equation}{equations}
\crefname{line}{line}{lines}
\crefname{section}{\S}{\S\S}
\Crefformat{section}{#2\S#1#3}
\crefrangeformat{section}{\S\S#3#1#4--#5#2#6}

\begin{document}

\maketitle

\begin{abstract}
Large language models (LLMs) have revolutionized numerous fields of research, driving significant advancements in natural language processing, machine translation, and beyond.
Although the extensive number of parameters contributes a lot to the great success, existing studies indicate that not all model parameters hold equal importance, which further leads to redundancy during the parameter update process.
Recent works for reducing redundant parameter updates for LLMs either lack task-specific data information, may leading to suboptimal model performance, or discard transformer components or insignificant parameters, limiting the model's scalability across different tasks and potentially compromising the LLM structure.
To address these issues and further enhance the performance of LLMs, we propose \methodfull{} (\method{}), a method that selectively updates parameters based on gradient information, which is specific to the target tasks.
Specifically, after calculating gradients during back propagation, we measure their absolute values and mask those with small absolute values.
Our empirical results in various training paradigms like SFT and DPO for various domains of tasks demonstrate that \method{} not only preserves the original network structure but also enhances the potential performance of LLMs.
Further analysis indicates that \method{} exhibits insensitivity to mask ratio and possesses computational efficiency comparable to vanilla training approach.
\end{abstract}

%

\section{Introduction}

Large language models (LLMs) have pivoted the centerpiece of various tasks such as textual understanding, natural language planning and instruction following \citep{achiam2023gpt, touvron2023llama, jiang2023mistral, wang2024inscl, luo2024chain}, which can be attributed to their extensive number of parameters, with performance and unique abilities significantly enhancing as the number of parameters increases.
To optimize the performance of LLMs in downstream tasks, full-parameter fine-tuning is commonly employed by researchers and practitioners.

However, a recent trend shows an observable phenomena that not all parameters of LLMs hold the same importance, which leads to redundancy during the parameter update process \citep{huang2024slim, jiang2024taia, yu2023language, luo2024velocitune}.
The existence of the redundancy of LLM parameter updates can be proven by several branches of works, like model pruning methods \citep{ma2023llm}, fine-tuning methods dropping the delta parameters \citep{yadav2024ties, yu2023language}, and so on, which shows minimal impact on performance.
Furthermore, inspired by recent works \citep{liu2024understanding, lv2024intelligent}, we believe that reducing redundancy in parameter updates has the potential to enhance model performance. \looseness=-1

Existing methods for reducing the redundancy of LLM parameter updates can be broadly classified into two categories based on their implementation. The first involves sparsifying the network during the training process using criteria such as randomness \citep{woo2024dropbp, hui2024hft}. However, a significant drawback of these methods is their failure to leverage information from task-specific data, may leading to suboptimal optimization objectives \citep{chen2024mofo}. The second focuses on eliminating non-essential structural components of the transformer block \citep{men2024shortgpt, song2024sleb} or discarding insignificant parameters \citep{yadav2024ties, yu2023language}. Yet, these one-off operations constrain the model's scalability across different tasks and impede its ability to adjust or reconfigure its computational complexity, which is crucial for adapting to varying task demands.

To address the above issues, a method that effectively identifies and retains the most critical LLM parameters during training is needed to reduce redundant updates and enhance overall model performance. 
In this work, we introduce the \methodfull{} (\method{}), a method that selects the set of parameters to be updated based on the gradient information associated with task-specific data.
Specifically, after calculating the gradients in the backward process, we measure the absolute values of gradients and mask those with small values.
There are several reasons for using gradient magnitude as a criterion.
Firstly, gradients inherently capture task-specific information and can be utilized without additional computational overhead during training.
Secondly, we assert that the gradient effectively assesses network parameter significance across different training datasets, allowing delicate control of the training process.
Thirdly, identifying and removing insignificant subsets of gradients during the training process, which does not have a drastic impact on the convergence of the model, and may even lead to a regularization effect that improves the performance of the model \citep{hoefler2021sparsity}. \looseness=-1

To evaluate the effectiveness of \method{}, we provide theoretical analysis and conduct experiments with  both Supervised Fine-tuning (SFT) and Direct Preference Optimization (DPO) training paradigms on several widely-used benchmarks on math reasoning, code generation, and general domain using various base models.
The experimental results demonstrate that \method{} exploits the gradient information to identify more important network parameters for sparse updating with acceptable extra time spent. This approach enhances model performance and reduces redundant parameter updates compared to several established baseline methods, including vanilla fine-tuning, Drop, and RMT.
Furthermore, it illustrates that \method{} is not sensitive to mask ratio selection and is more effective than random parameter updating. We further analyze the computational FLOPs and time efficiency of GMT, demonstrating that it enables a plug-and-play replacement of vanilla SFT without destroying the network architecture. \looseness=-1

Our contributions can be summarized as:
\begin{itemize}[itemsep=0pt, topsep=4pt]  
\item We propose \method{}, which masks the gradients with small absolute values and discriminates the importance of parameters during training, naturally utilizing the information of task-specific data. This approach effectively reduces redundant parameter updates to the LLM and enhances performance on downstream tasks.
\item We conduct both theoretical analysis and exhaustive experiments with different base models on several benchmarks comparing with representative baselines. The results confirm the effectiveness of \method{}. 
\item We further demonstrate the adaptability, robustness and efficiency of the \method{} by analyzing the drop strategy, mask ratio and time efficiency. 
\end{itemize}

\section{Related Works}
Reducing redundant parameter updates in LLMs is a significant research problem \citep{dalvi2020analyzing}. Addressing this issue can prevent the waste of computational resources and optmize the model training process.
We categorized the reduction of redundant parameter updates into post-processing and in-training based on the stage of implementation. This section describes the properties of the typical methods of both strategies as well as the respective  drawbacks.

\paragraph{Post-processing Strategy}
The post-processing strategy, although requiring tuning with full parameter participation, achieves the same purpose as in-training through a series of tuned parameter drop and reset strategies.
Inspired by dropout, mixout \citep{lee2019mixout} stochastically mixes source and target parameters as a regularization technique and extends to fine-tune downstream tasks.
DARE \citep{yu2023language} finds that setting most (90\% or even 99\%) of the delta parameters to zero does not diminish the ability to fine-tune the LLM, and extends to model fusion accordingly.
Ties-merging \citep{yadav2024ties} describes the delta parameter as a type of task vector that trims it, and then averages the parameters that are identical for multiple model aggregation symbols to achieve model merging.
ExPO \citep{zheng2024weak} is inspired by model interpolation \citep{wortsman2022model} and obtains better aligned models by direct extrapolation from the parameters of the aligned model obtained by DPO or RLHF and the initial SFT model.
Although these methods can be implemented directly on fine-tuned models that are open access and require little additional computational overhead, they never achieve optimal performance due to the lack of refinement in the training process. 
Some methods attempt to directly remove redundant layers \citep{men2024shortgpt} or components of transformer blocks \citep{song2024sleb} in LLMs based on quantitative criteria. However, these methods compromise the structural integrity of LLMs, hindering their applicability across a wider range of scenarios. \looseness=-1

\paragraph{In-training Strategy} 
The in-training strategies aim to dynamically mitigate the impact of model parameter redundancy during the training process. DropBP \citep{woo2024dropbp} calculates the sensitivity of each layer to allocate an appropriate dropout rate, thereby randomly dropping certain layers during backpropagation to improve efficiency and reduce redundancy. HFT \citep{hui2024hft} randomly selects half of parameters for learning new tasks while freezing the remaining parameters to preserve previously learned knowledge, thus eliminating the redundancy of parameter knowledge across different learning stages. PAFT \citep{pentyala2024paft} separately conducts the SFT and DPO training processes, using L1 regularization to sparsify delta parameters and reduce redundancy, thereby achieving alignment and parallel training. These methods either employ randomness or regularization during the training process and lack guidance from task-specific data.
Our \method{} is focused on the task-specific data information and fine-grained tuning of parameters, which not only preserves the model’s generalizability but also significantly improves its adaptability and performance on specific tasks.

\section{Methodology}

\begin{figure*}[t]
\centering
\includegraphics[width=1\linewidth]{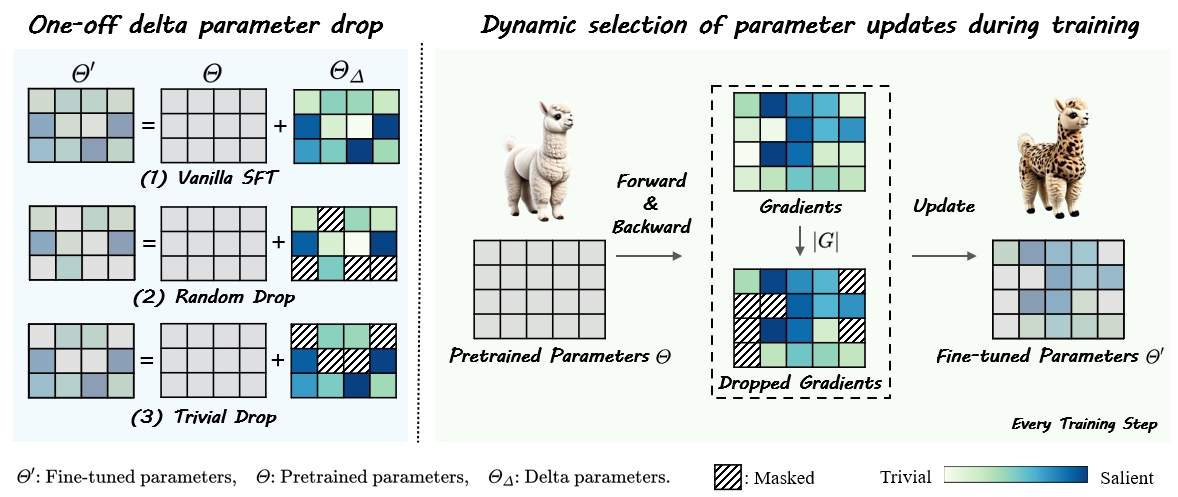}
\caption{Illustration of our proposed method \method{}, compared with the one-off drop delta parameters approach. The figure on the left delineates the distinction between a trivial drop and a random drop, wherein the trivial drop serves to diminish the redundant updates that arises during the fine-tuning process. Building upon this insight, we refine the training procedure by preferentially updating more significant parameters, as determined by the gradient information pertinent to the task-specific data. This selective updating is operationalized through the implementation of a masking strategy that filters out gradients with smaller absolute values. 
}
\label{fig:framework}
\end{figure*}

In this section, we elucidate the principle underlying the \method{} method, providing a comprehensive explanation through mathematical formulation and theoretical analysis.  \cref{fig:framework} presents the framework of our method and compares it with a one-off drop strategy for delta parameters (i.e., the difference between the fine-tuned parameters and the pre-trained parameters). Specifically, \method{} effectively identifies the most critical parameters to be updated during training by the absolute magnitude of the gradient. This dynamic selection strategy prevents redundant parameter updates, thereby improving the performance of LLM.

\subsection{\methodfull{}}
\label{sec:method}

To decrease redundant updates by leveraging the information of task-specific data during training, while preserving the integrity of the original LLM architecture, we propose \method{} for selecting critical parameters.

For each batch of training data, we compute the gradient of the loss function $\mathcal{L}(\Theta)$ with respect to the parameters. To facilitate gradient accumulation, we define an accumulation interval \(N\), representing the number of mini-batches over which gradients are accumulated before performing a parameter update. The accumulated gradient \(\Gamma_{ij}\) for each parameter \(\theta_{ij}\) is defined as:

\begin{small}
\begin{equation}
\Gamma_{ij} = \frac{1}{N} \sum_{n=1}^{N} \nabla_{\theta_{ij}} \mathcal{L}(\Theta, \mathcal{B}_n) 
\end{equation}
\end{small}
where \(\mathcal{B}_n\) represents the \(n\)-th mini-batch of data and \(\nabla_{\theta_{ij}} \mathcal{L}(\Theta, \mathcal{B}_n)\) is the gradient of the loss function with respect to \(\theta_{ij}\) for the mini-batch.  
   
Upon accumulating the gradients over \(N\) mini-batches, we employ gradient information as a signal to identify the importance of parameters at the element-wise level in a fine-grained manner.
This operation involves sorting the components of each accumulated gradient \(\Gamma_{ij}\) by their absolute values and selecting a pre-defined top percentile \(k\) for updating the parameters. The masked gradient \(\mathcal{M}(\Gamma_{ij}, k)\) is calculated as: 

\begin{small}
\begin{equation}
\mathcal{M}(\Gamma_{ij}, k) = \left\{ g_{ij} \ | \ g_{ij} \in \Gamma_{ij}, \ |g_{ij}| \geq T_k \right\} 
\end{equation}
\end{small}
where \(g_{ij}\) represents the \(ij\)-th component of the accumulated gradient for parameter \(\theta_{ij}\). The threshold \(T_k\) is the value such that the absolute values of the components of \(\Gamma_{ij}\) that are greater than or equal to \(T_k\) fall within the top \(k\) percentile of all components by magnitude.  
   
The subsequent parameter update step utilizes the masked gradient \(\mathcal{M}(\Gamma_{ij}, k)\):

\begin{small}
\begin{equation}
\theta_{ij}^{(t+1)} = \theta_{ij}^{(t)} - \eta \cdot \mathcal{M}(\Gamma_{ij}, k)
\end{equation}
\end{small}
where \(\eta\) is the learning rate and \(t\) indexes the current training step. The detailed algorithmic training procedure is given in \cref{alg:sparse_fine_tuning_with_gradmask}. 

\begin{algorithm}[tb]
\caption{\methodfull{} }
\label{alg:sparse_fine_tuning_with_gradmask}
\textbf{Hyperparameters}: The accumulation interval $N$, the training step $T$ and the learning rate $\eta$. \\
\textbf{Input}: The initial model parameters $\Theta^{(0)}$ and the \(n\)-th mini-batch of data \(\mathcal{B}_n\). \\
\textbf{Output}: The updated model parameters $\Theta^{(T)}$. 

\begin{algorithmic}[1]
\FOR{$t$ in $0 \to {T-1}$}
\STATE $\Gamma \gets 0$
\FOR{$n$ in $1 \to N$}
\STATE $\Gamma \gets \Gamma + \nabla_{\Theta} \mathcal{L}(\Theta^{(t)}, \mathcal{B}_n)$
\ENDFOR
\STATE $\Gamma \gets \frac{1}{N} \Gamma$
\STATE $T_k \gets \text{Percentile \ threshold \ based \ on \ } \Gamma$
\STATE $\mathcal{M}(\Gamma, k) \gets \left\{ g_{ij} \ | \ g_{ij} \in \Gamma_{ij}, \ |g_{ij}| \geq T_k \right\}$  
\FOR{$\theta_{ij} \in \Theta$}
\IF{$|\Gamma_{ij}| \geq T_k$}
\STATE $\theta^{(t+1)}_{ij} \gets \theta^{(t)}_{ij} - \eta \cdot \Gamma_{ij}$
\ENDIF
\ENDFOR
\ENDFOR
\end{algorithmic}
\end{algorithm}

\subsection{Theoretical Analysis}
\label{sec:slct_tune_para}
We present the theoretical analysis in two parts.
First, we explain from the optimization perspective \citep{fu2023effectiveness, hui2024hft} why removing unimportant parameter updates can enhance model performance, by fine-tuning partial parameters  to optimize the upper bound of the original loss function in Appendix.
Then, we discuss the theoretical analysis of the absolute value of the gradient used in \method{} , demonstrating that this absolute value can effectively signify the importance of the parameter.

In order to consider the impact of task-specific data on the importance of each parameter in a model, we identify and update only a sparse subset of parameters based on their impact on the loss function. We define the loss function $L(\Theta; \mathcal{D})$ over the dataset $\mathcal{D}$ as:

\begin{small}
\begin{equation}
\mathcal{L}(\Theta; \mathcal{D}) = \frac{1}{n} \sum_{i=1}^n \ell(\Theta; (\mathbf{x}_i, y_i))
\end{equation}
\end{small}
where $\ell$ represents the loss for a single data point, with $\mathbf{x}_i$ and $y_i$ denoting the input features and corresponding label, respectively. The full parameter set is denoted by $\Theta$.

To discern the impact of individual parameters $\theta_{ij}$ on the loss function $\mathcal{L}(\Theta; \mathcal{D})$,  we consider the impact of their removal while keeping all other parameters constant. 
The differential effect of excluding a parameter $\theta_{ij}$ is quantified by the change in loss $\Delta \mathcal{L}_{ij}(\Theta; \mathcal{D})$, which is formulated as:

\begin{small}
\begin{equation}
\Delta \mathcal{L}_{ij}(\Theta; \mathcal{D}) = \mathcal{L}(\mathbf{I} \odot \Theta; \mathcal{D}) - \mathcal{L}((\mathbf{I} - \mathcal{E}_{ij}) \odot \Theta; \mathcal{D})
\end{equation}
\end{small}
where $\mathbf{I}$ represents the identity matrix and $\mathcal{E}_{ij}$ denotes an indicator matrix that has the same dimensions as $\Theta$. In $\mathcal{E}_{ij}$, all elements are zero except for the $(i, j)$ element, which is one. The Hadamard product, indicated by $\odot$, performs an element-wise multiplication, isolating the effect of the single parameter $\theta_{ij}$. 

Given the computational infeasibility of evaluating $\Delta \mathcal{L}_{ij}$ for each parameter, we invoke the first-order Taylor series expansion around the current parameter vector $\Theta$, which provides a linear approximation of the loss function's behavior in the vicinity of $\Theta$. The first-order approximation is represented by the gradient of the loss function with respect to $\theta_{ij}$:

\begin{small}
\begin{equation}
\Delta \mathcal{L}_{ij}(\Theta; \mathcal{D}) \approx \nabla_{\theta_{ij}} \mathcal{L}(\Theta; \mathcal{D}) \cdot (-\theta_{ij})
\end{equation}
\end{small}
where $\nabla_{\theta_{ij}} \mathcal{L}(\Theta; \mathcal{D})$ denotes the partial derivative of the loss function concerning the parameter $\theta_{ij}$. The negative sign arises from the fact that we are considering the removal of the parameter, which corresponds to a negative perturbation in its value.

Our objective is to discern the important parameters within the network architecture that are relevant to task-specific data. To this end, we employ the magnitude of the gradient $\nabla_{\theta_{ij}}$ as the criterion for saliency. It is imperative to recognize that a high magnitude of the gradient (regardless of its sign) typically denotes that the parameter $\theta_{ij}$ exerts a substantial influence on the loss function, whether the effect is positive or negative. Consequently, such parameters must be preserved to facilitate learning on the corresponding weights.
The absolute value of the gradient $\nabla_{\theta_{ij}}$ is then used as a saliency measure to determine the importance of the parameter:

\begin{small}
\begin{equation}
 s_{ij} = \left| \nabla_{\theta_{ij}} \mathcal{L}(\Theta; \mathcal{D}) \right|
\end{equation}
\end{small}
Using this saliency measure, we construct the binary mask matrix $M$ such that $M_{ij} = 1$ if the parameter $\theta_{ij}$ is deemed important based on a predefined sparsity level $\kappa$:

\begin{small}
\begin{equation}
M_{ij} = \mathbb{I}[s_{ij} \ge \tilde{s}_{(\kappa)}]
\end{equation}
\end{small}
where $\tilde{s}_{(\kappa)}$ is the saliency threshold selected to ensure that only the top $\kappa$ influential parameters have their corresponding entries in $M$ set to one. The indicator function $\mathbb{I}[\cdot]$ yields one if the condition within the brackets is true and zero otherwise.

\section{Experiments}
\subsection{Experimental Setup}

To evaluate the effectiveness of the \method{} approach, we conducted a comprehensive assessment of various models across a range of tasks, including code generation, mathematical reasoning, and general domains. Our experiments encompassed two training paradigms, SFT and DPO, allowing for a comparative analysis of the baseline performance in diverse settings. Furthermore, we performed further experimental analysis of mask ratio, drop strategy, and the overall efficiency of the approach.

\paragraph{Training and Evaluation.}

For code generation, we employ the Magicoder-Evol-Instruct-110K \citep{wei2023magicoder} as the training data, which is a decontaminated version of evol-codealpaca-v1\footnote{\url{https://huggingface.co/datasets/theblackcat102/evol-codealpaca-v1}}. We utilize \textsc{Mistral-7B} ~\citep{jiang2023mistral} and \textsc{DeepSeek-Coder-Base-6.7B}~\citep{guo2024deepseek} as the base models.
The trained models are evaluated using the HumanEval~\citep{chen2021evaluating} and MBPP~\citep{austin2021program} benchmarks, which are widely recognized for their effectiveness in measuring the proficiency of Python text-to-code generation.
To enable a more comprehensive evaluation, we also introduce HumanEval+ and MBPP+, both of which are provided by EvalPlus \footnote{\url{https://evalplus.github.io/leaderboard.html}} ~\citep{liu2024your}. 
For math reasoning, the MetaMathQA ~\citep{yu2023metamath} dataset is employed to fine-tune on the \textsc{Mistral-7B} and \textsc{LLaMA3}-8B models.
The evaluation is conducted using the GSM8k~\citep{cobbe2021training} and MATH~\citep{hendrycks2021measuring} benchmarks, which are specifically constructed to test the model's capacity for mathematical reasoning and problem-solving.
For the general domain, the \textsc{T\"ulu}~V2 ~\citep{wang2024far} dataset is utilized in SFT phase training on the \textsc{LLaMA2}-7B ~\citep{touvron2023llama} and \textsc{LLaMA2}-13B model, the UltraFeedback \citep{cui2023ultrafeedback} is utilized in DPO phase training.
Following HFT~\citep{hui2024hft}, we evaluate model on MMLU~\citep{hendrycks2020measuring}, GSM8k~\citep{cobbe2021training}, BBH~\citep{suzgun2023challenging}, TyDiQA~\citep{clark2020tydi},  TruthfulQA~\citep{lin2022truthfulqa} and HumanEval~\citep{chen2021evaluating}.

\begin{table*}[!htbp]
\centering
\begin{tabular}{ccccccccc}
\toprule
\multicolumn{1}{c|}{\textbf{Method}} & \textbf{HumanEval} & \textbf{HumanEval+} & \textbf{MBPP} & \textbf{MBPP+} & \multicolumn{1}{c|}{\textbf{Average}} & \textbf{GSM8k} & \textbf{MATH} & \multicolumn{1}{c}{\textbf{Average}} \\  \midrule 

\multicolumn{1}{c|}{} & \multicolumn{5}{c|}{\footnotesize \textsc{Mistral-7B}}                              & \multicolumn{3}{c}{\footnotesize  \textsc{Mistral-7B}} \\   \midrule
\multicolumn{1}{c|}{Pre-trained$^\dagger$} & 28.7 & 23.8 & 51.9 & 42.1 & \multicolumn{1}{c|}{36.6}  & -            & -           & -         \\
\multicolumn{1}{c|}{SFT}         & 68.3 & 64.0 & 56.1 & 46.9 & \multicolumn{1}{c|}{58.8} & 75.3               & 27.0   & 51.2          \\
\multicolumn{1}{c|}{Drop} & 68.3  & 64.0 & 54.6  & 46.1 & \multicolumn{1}{c|}{58.3} & 74.6               & 27.6      & 51.1              \\

\multicolumn{1}{c|}{HFT}         & 67.1 & 61.6 & 57.1 & 48.1 & \multicolumn{1}{c|}{58.5} & 77.8               & 27.3      & 52.6        \\
\multicolumn{1}{c|}{RMT} & \textbf{70.7} & \textbf{64.6} & 55.1 & 45.4 & \multicolumn{1}{c|}{59.0} & 74.6               & 25.1    &   49.9       \\
\multicolumn{1}{c|}{{\method{}}}    & {69.5} & {62.2} & {\textbf{59.6}} & {\textbf{48.6}} & \multicolumn{1}{c|}{{\textbf{60.0}}} & {\textbf{78.6}}       & {\textbf{28.5}}     &     {\textbf{53.6}}     \\ \midrule
\multicolumn{1}{c|}{} & \multicolumn{5}{c|}{\footnotesize \textsc{DeepSeek-Coder-Base-6.7B}}                              & \multicolumn{3}{c}{\footnotesize \textsc{Llama3-8B}} \\ \midrule
\multicolumn{1}{c|}{Pre-trained$^\dagger$} & 47.6 & 39.6 & 72.0 & 58.7 & \multicolumn{1}{c|}{54.5} & -                  & -        & -               \\
\multicolumn{1}{c|}{SFT}         & 76.8 & 73.8 & 74.9 & 62.4 & \multicolumn{1}{c|}{72.0} & 78.1               & 29.2       & 53.7       \\
\multicolumn{1}{c|}{Drop} & 76.2  & 72.6 & 74.7  & 62.4  & \multicolumn{1}{c|}{71.5} & 78.1               & 29.4      & 53.8             \\

\multicolumn{1}{c|}{HFT}         & 74.4 & 70.1 & 75.2 & 62.9 & \multicolumn{1}{c|}{70.7} & 74.2               & 28.5      & 51.4        \\
\multicolumn{1}{c|}{RMT} & 72.6 & 67.1 & \textbf{78.8} & \textbf{67.5} & \multicolumn{1}{c|}{71.5} & 80.3               & 31.3       &55.8        \\
\multicolumn{1}{c|}{{\method{}}}    & {\textbf{78.0}} & {\textbf{74.4}} & {75.7} & {63.7} & \multicolumn{1}{c|}{{\textbf{73.0}}} & {\textbf{82.0}}           & {\textbf{32.0}}     & {\textbf{57.0}}           \\ \bottomrule
\end{tabular}%
\caption{Experimental results for a single task in a specific domain. All models are evaluated with zero-shot prompting. $^\dagger$ We obtain the results of the code benchmark from EvalPlus. ``-'' denotes that no zero-shot results were officially reported. Bold text indicates the best results for the fine-tuned model on each benchmark.}
\label{tab:main_math&code}
\end{table*}

\paragraph{Implementation Details.}

We choose different base models for different tasks.
All training experiments were done on NVIDIA A100 and NVIDIA H100 machines.
In addition, we utilize BFloat16 precision and set the weight decay to 0.
We use the cosine learning rate scheduler after a linear warm-up stage with a ratio of 0.03.

\paragraph{Baselines.}
In order to thoroughly evaluate the effectiveness of our method, we compare \method{} to the following baselines:
\begin{itemize}[itemsep=0pt, topsep=4pt]

\item \textbf{SFT}: Vanilla supervised fine-tuning.

\item \textbf{Drop}: As an extension of \citep{yu2023language}, dropping a preset ratio of trivial delta parameters on a one-off basis after the vanilla supervised fine-tuning.

\item \textbf{HFT}: Half Fine-Tuning,  half of the parameters are selected for learning the new task, and the other half is frozen to retain previous knowledge.

\item \textbf{RMT}: Random Mask-Tuning, a preset ratio of the parameters are randomly updated at a fine-grained element-wise level during the training process.
\end{itemize}

To ensure fairness, we apply an identical mask ratio across Drop, RMT, and \method{}.

\subsection{Main Results}

\paragraph{Code Generation.} The experimental results for the code generation task are presented in \cref{tab:main_math&code}.
By integrating task-specific gradient information with avoidance of random parameter selection, \method{} achieves an average performance improvement of 1.2\% on the \textsc{Mistral-7B} model and 1\% on the \textsc{DeepSeek-Coder-Base-6.7B} model. 
In contrast, HFT freezes half of the parameters to maintain the original capabilities of the model, and the learning process is hampered by the lack of sufficient parameter updates for a specific task domain.
Furthermore, the RMT method shows some benefits in reducing redundant parameter updates on \textsc{Mistral}-7B. However, it fails to maintain this performance on another model, suggesting that it lacks robustness due to its inability to utilize task-specific data information.
Compared to SFT, a one-off drop directly on a fine-tuned model does not improve performance due to its sensitivity to the optimal drop ratio.
In the specialized field of code generation, this improvement underscores the efficacy of \method{} in optimizing model performance through fine-grained parameter selection. \looseness=-1

\paragraph{Math Reasoning.}
\cref{tab:main_math&code} presents a comparative analysis of the \method{} method against various baseline approaches.
\method{} outperforms other methods by leveraging gradient information to capitalize on sparsity independent of the inconsistencies associated with random methods.
Particularly, \method{} performs excellently on the GSM8k benchmark, outperforming SFT by 3.3\% and 3.9\% on both models.
In contrast, the HFT method shows an improvement over the baseline on \textsc{Mistral-7B}. However, it performs terribly on \textsc{Llama3-8B}, showing the instability of the method, which is attributed to the strategy of randomly chosen parameters.
Drop is slightly improved on the MATH benchmark but not as effective as the \method{} method that performs the drop operation throughout the training process.
In summary, this strategic use of gradient signal associated with the task-specific data ensures stable and progressive performance improvements throughout the learning process, as evidenced by its success across different models.

\paragraph{General Domain.}

\begin{table*}[th]
\centering
\small                 
\begin{tabular}{lcccccccc}
\toprule
\multicolumn{1}{c}{\multirow{2}{*}{\textbf{Model}}} & \multicolumn{1}{c}{\multirow{2}{*}{\textbf{Method}}} & \textbf{MMLU} & \textbf{GSM8k} & \textbf{BBH} & \textbf{TyDiQA} & \textbf{TruthfulQA} & \textbf{HumanEval} & \multirow{2}{*}{\textbf{Average}} \\
 &  & \scriptsize0-shot, EM & \scriptsize8-shot CoT, EM & \scriptsize3-shot CoT, EM & \scriptsize1-shot, F1 & \scriptsize0-shot, MC2 & \scriptsize0-shot, Pass@10 &  \\ \midrule
\textsc{Llama2-7B} & Pre-trained & 41.6 & 12.0 & 39.9 & 48.4 & 38.5 & 26.2 & 34.4 \\
\textsc{Llama2-13B} & Pre-trained & 52.2 & 34.5 & 50.7 & 50.3 & 49.8 & 32.7 & 45.0 \\ \midrule
\multicolumn{9}{c}{\texttt{Supervised Fine-tuning (SFT) on \textsc{T\"ulu}~V2}} \\ \midrule
\multirow{4}{*}{\textsc{Llama2-7B}} & SFT & 48.5 & 25.0 & 42.2 & 51.2 & 41.7 & \textbf{36.9} & 41.0 \\
 & HFT & \textbf{50.8} & 30.5 & 43.6 & 52.3 & 45.4 & 34.6 & 42.9 \\
 & RMT & 47.4 & 34.5 & \textbf{44.4} & 52.9 & \textbf{47.9} & 33.7 & 43.4 \\
 & {\method{}} & 47.6 & \textbf{38.0} & 43.3 & \textbf{53.1} & 47.5 & 34.6 & \textbf{44.0} \\  \midrule
\multirow{4}{*}{\textsc{Llama2-13B}} & SFT & 50.6 & 45.0 & 47.8 & 55.0 & 42.6 & 42.4 & 47.2 \\
 & HFT & 54.5 & 46.5 & \textbf{53.7} & 56.7 & 45.7 & \textbf{43.5} & 50.1 \\
 & RMT & 54.6 & 50.5 & 52.8 & 56.5 & 45.2 & 41.4 & 50.2 \\
 & {\method{}} & \textbf{54.6} & \textbf{54.0} & 51.5 & \textbf{57.1} & \textbf{46.0} & 39.5 & \textbf{50.5} \\ \midrule
\multicolumn{9}{c}{\texttt{Direct Preference Optimization (DPO) on UltraFeedback}} \\ \midrule
\multicolumn{1}{l}{\multirow{4}{*}{\textsc{Llama2-7B}}} & DPO & \multicolumn{1}{c}{48.9} & \multicolumn{1}{c}{28.0} & \multicolumn{1}{c}{42.9} & \multicolumn{1}{c}{50.2} & \multicolumn{1}{c}{45.7} & \multicolumn{1}{c}{35.6} & \multicolumn{1}{c}{41.9} \\
\multicolumn{1}{c}{} & HFT &\multicolumn{1}{c}{ \textbf{50.7}} & \multicolumn{1}{c}{30.5} & \multicolumn{1}{c}{42.8} & \multicolumn{1}{c}{43.9} & \multicolumn{1}{c}{49.8} & \multicolumn{1}{c}{35.1} & \multicolumn{1}{c}{42.1} \\
\multicolumn{1}{c}{} & RMT & \multicolumn{1}{c}{48.5} & \multicolumn{1}{c}{34.0} & \multicolumn{1}{c}{43.8} & \multicolumn{1}{c}{53.1} & \multicolumn{1}{c}{54.7} & \multicolumn{1}{c}{37.1} & \multicolumn{1}{c}{45.2} \\
\multicolumn{1}{c}{} & {\method{}} & 48.5 & \textbf{36.0} & \textbf{44.7} & \textbf{53.8} & \textbf{54.7} & \textbf{39.8} & \textbf{46.3} \\ \midrule
\multicolumn{1}{l}{\multirow{4}{*}{\textsc{Llama2-13B}}} & DPO & \multicolumn{1}{c}{52.0} & \multicolumn{1}{c}{44.0} & \multicolumn{1}{c}{47.1} & \multicolumn{1}{c}{51.5} & \multicolumn{1}{c}{45.5} & \multicolumn{1}{c}{\textbf{44.3}} & \multicolumn{1}{c}{47.4} \\
\multicolumn{1}{c}{} & HFT & \multicolumn{1}{c}{\textbf{55.0}} & \multicolumn{1}{c}{45.5} & \multicolumn{1}{c}{51.4} & \multicolumn{1}{c}{53.2} & \multicolumn{1}{c}{49.5} & \multicolumn{1}{c}{42.9} & \multicolumn{1}{c}{49.6} \\
\multicolumn{1}{c}{} & RMT & \multicolumn{1}{c}{54.4} & \multicolumn{1}{c}{55.5} & \multicolumn{1}{c}{50.9} & \multicolumn{1}{c}{56.3} & \multicolumn{1}{c}{51.7} & \multicolumn{1}{c}{40.5} & \multicolumn{1}{c}{51.6} \\
\multicolumn{1}{c}{} & {GMT} & 54.5 & \textbf{56.0} & \textbf{51.9} & \textbf{56.7} & \textbf{53.3} & 41.1 & \textbf{52.3} \\ \bottomrule
\end{tabular}
\caption{Results of \textsc{Llama2-7B} and \textsc{Llama2-13B} fine-tuned on the \textsc{T\"ulu}~V2 dataset. The results of Pre-trained, SFT, and HFT are taken from \citep{hui2024hft}. Bold text indicates the best results on each benchmark. The DPO phase is initialized with the corresponding methods using the HFT- and GMT-based SFT models fine-tuned on \textsc{T\"ulu}~V2, respectively. }
\label{tab:main_tulu}
\end{table*}

\cref{tab:main_tulu} shows the experimental comparison of the proposed \method{} with the baseline.
In the experiments with the fine-tuned \textsc{Llama2}-7B model, the \method{} method outperforms the SFT by 3.0\% on average across all tasks, and in particular, it significantly leads by 13.0\% on GSM8k. 
\method{} likewise demonstrated superior performance than HFT in multitasking scenarios in the general domain, obtaining an average lead of 1.1\%.
After the expansion of the LLM size, the \method{} maintains its performance, with a 3.3\% improvement in fine-tuning performance of \textsc{Llama2}-13B compared to the vanilla SFT.
Notably, both RMT and HFT perform better than SFT, suggesting that the general domain of multi-tasking benefits from appropriate sparsity, even when the parameter selection strategy is completely random.
With the same amount of parameter updates, \method{} shows further improvement over RMT, suggesting that the strategy of selecting parameters based on gradient information derived from task-specific data in our method is practically effective.
After extending the \method{} to DPO, the superiority of the GMT approach remains evident. Experimental results across two model sizes indicate that \method{} identifies parameters more worthy of updating in multi-task learning, thereby reducing the redundancy of parameter updates and enhancing model performance.

\subsection{Further Discussions}
\paragraph{Analysis of Mask Ratio.} 

\begin{figure*}[h]
    \centering
    \includegraphics[width=0.9\linewidth]{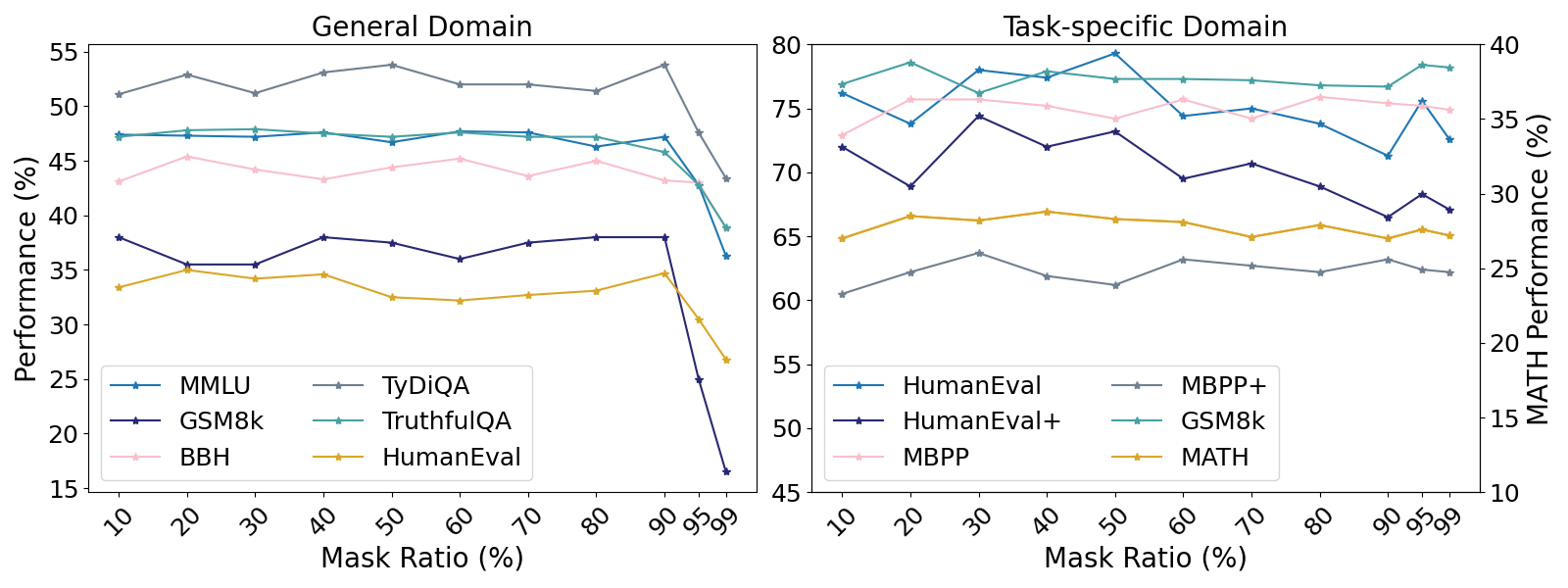}
    \caption{Fine-tuning performance of the proposed GMT with respect to various mask ratios during training in three domains.
    The training LLMs utilized for code generation task, math reasoning task, and general domain are \textsc{DeepSeek-Coder-Base-6.7B}, \textsc{Mistral-7B}, and \textsc{Llama2-7B}, respectively.
    The experimental results for the MATH benchmark are presented on the secondary y-axis located on the right side of the figure.}
    \label{fig:mask_ratio}
\end{figure*}

To comprehensively ascertain the sensitivity of the hyperparameter mask ratio selection on \method{}, we analyze the influence of the amount of parameter updates during training. Comparative experiments were conducted across three domains. 
We iterated through all experiments, applying the range of parameters to be fine-tuned with 10\% granularity. For extreme cases, experiments were performed with mask ratios of 95\% and 99\%. 
As shown in \cref{fig:mask_ratio}, the experimental results indicate that despite fluctuations in the number of parameter updates,  the proposed method maintains a relatively consistent performance level with no more than 90\% mask ratio. 
Our \method{} reveals that the optimal performance in fields like mathematics and coding, which require specialized knowledge, is achieved with mask ratios of 20\% to 40\%.
Even under extreme mask ratio settings, \method{} maintains impressive performance in task-specific domains such as math reasoning and code generation. However, its effectiveness significantly diminishes in the general domain within multi-task scenarios. This suggests a stronger adaptability of \method{} to single-task scenarios, where the entire learning process can be completed with merely a 1\% parameter update. Conversely, in multi-task settings, the potential lack of sufficient parameter updates may lead to an inadequate learning process. Further analysis demonstrates that the performance of the proposed \method{} exhibits robustness to variations in the mask ratio, despite the latter being an adjustable hyperparameter within the model's configuration. \looseness=-1

\paragraph{Analysis of Drop Strategy.}

Our analysis reveals that not all delta parameters contribute equally to the model's performance. Specifically, we find that parameters with salient values are critical for maintaining the efficacy of the LLM, and their removal leads to a notable and rapid degradation in model functionality.  
To systematically explore this phenomenon, we designed a series of experiments using the \textsc{Mistral-7B} model, focusing on domain-specific applications in mathematics (GSM8k and MATH).
The experimental results are depicted in Appendix.   \looseness=-1
   
Our results indicate substantial differences in the impact of each strategy on model performance. The strategy of dropping a portion of the delta parameters with small absolute values after vanilla fine-tuning demonstrates a relatively robust performance, only showing significant degradation when the dropout rate reaches 80\%.  In contrast, the random drop strategy leads to an apparent performance decline at a much lower sparsity level of 40\%.   
The strategy of dropping parameters with salient magnitudes results in a precipitous decline in model capabilities. This finding underscores the importance of these significant delta parameters in sustaining the functional integrity of the model. 
Expectedly, the utilization of the gradient information to trivial drop during the training process can elevate the upper limit of LLM performance and is not constrained by drop rate. \looseness=-1
   
Such experiments demonstrate that the parameter updates during the fine-tuning process of LLM are redundant, and that the use of a reasonable sparse updating strategy enables the model to learn better.  Comparison with the experiments of random drop indicates that our strategy of considering task-specific data during training, using gradient information as a signal, and retaining  neurons with large update magnitudes for completing the update is effective. \looseness=-1

\begin{table}[t]
\small
\centering
\begin{tabular}{cccc}
\toprule
\multirow{2}{*}{\textbf{Model}}      & \multirow{2}{*}{\textbf{Method}} & \textbf{Traning Speed} & \textbf{FLOPs}                    \\
                       &                         & (samples/s) & ($1e^{18}$) \\  \midrule
\multirow{3}{*}{\begin{tabular}[c]{@{}c@{}} \textsc{Mistral-7B} \\ Code Generation \end{tabular}} & Vanilla SFT             & 15.32         & 1.481                    \\
                       & RMT             & 13.37         & 1.481                    \\
                       & \method{}                     & 13.43         & 1.480                    \\  \midrule
\multirow{3}{*}{\begin{tabular}[c]{@{}c@{}} \textsc{Llama2-13B} \\ General Domain \end{tabular}} & Vanilla SFT                     & 13.53         & 1.706                    \\
                       & RMT                    & 12.54         & 1.707                    \\
                       & \method{}                     & 12.74         & 1.707                    \\            
\bottomrule
\end{tabular}%
\caption{Comparison of training speed and computational FLOPs, the number of train samples per second is an indicator of training speed.}
\label{tab:anal_effi}
\end{table}

\paragraph{Analysis of Efficiency.} 
To demonstrate the time-efficiency of proposed \method{},
we compare the training speed and computed FLOPs of \method{}, vanilla SFT, and RMT, with the metric of number of training samples per second responding to the training speed, and the results are shown in \cref{tab:anal_effi}. The FLOPs computed by the three tuning strategies during training are quite comparable due to the fact that the gradient information utilized by \method{} is a by-product of model training and does not impose additional derivation operations.
In terms of training speed, the RMT is close to GMT, being 14\% and 6\% slower than vanilla SFT in the code generation and general domain, respectively. 
Since GMT needs to calculate the threshold of the gradient being masked based on a preset ratio, 
a process that requires a \lstinline|TopK| operation after the model backpropagates the gradient, and then drops values with smaller absolute values of the gradient based on the threshold. 
Nonetheless, this trade-off between a small minor time overhead and performance improvement is completely acceptable. \looseness=-1

\section{Conclusion}
In this paper, we propose the \methodfull{} (\method{}), a pragmatic approach to optimize LLMs by selectively updating parameters based on gradient information.
\method{} identifies more important parameters to eliminate redundant updates introduced in fine-tuning and improves model performance on various tasks.
We evaluate the performance of \method{} on multiple models in the code generation, math reasoning, and general domain. Experiments demonstrate that \method{} not only elevates the upper limits of LLM performance across diverse tasks but also exhibits robustness to various fine-tuning settings. We further analyze the computational efficiency of \method{} to confirm that no additional computational FLOPs are needed, as well as acceptable extra time consumption.
Moreover, \method{} allows for fine-tuning without destroying the network structure, and thus can readily replace SFT and DPO to accomplish the optimization process.

\bibliography{aaai25}

\appendix
\section{Effectiveness of Sparsity in Fine-tuning}
\label{sec:eff_SparseFT}
We consider the task of minimizing the loss function \(\mathcal{L}(\Theta)\) for a large language model (LLM) with parameters \(\Theta\), subject to a sparsity constraint on the parameter updates. This can be formalized as the following optimization problem:

\begin{equation}
\begin{gathered}
\min_\Theta \mathcal{L}(\Theta) \\
\text{subject to} \quad \|(I - M)\Theta_\Delta\|^2 = 0,
\end{gathered}
\label{eqn:optim_problem}
\end{equation}
 where $\mathbf{I}$ represents the identity matrix, ensuring that the constraint applies across all dimensions of \(\Theta\).

By introducing a vector of Lagrange multipliers \(\lambda\), we construct the Lagrangian function \(\mathcal{L}_d\) associated with our optimization problem (\ref{eqn:optim_problem}):

\begin{equation}
\mathcal{L}_d(\Theta, \lambda) = \mathcal{L}(\Theta) + \lambda^\top \|(\mathbf{I} - M)\Theta_\Delta\|^2.
\end{equation}

The dual problem seeks to find the saddle point of \(\mathcal{L}_d\) through a min-max strategy:

\begin{equation}
\bar{\mathcal{L}} = \max_\lambda \min_\Theta \mathcal{L}_d(\Theta, \lambda).
\label{eqn:dual_problem}
\end{equation}

The min-max inequality allows us to establish an upper bound for our primal problem. For the dual problem (\ref{eqn:dual_problem}), the min-max inequality suggests:

\begin{equation}
\begin{aligned}
& \min_\Theta \mathcal{L}(\Theta) + \|(\mathbf{I} - M)\Theta_\Delta\|^2   \\
& \leq \max_\lambda \min_\Theta \mathcal{L}_d(\Theta, \lambda)  \\
& \leq \min_\Theta \max_\lambda \mathcal{L}_d(\Theta, \lambda).
\label{eqn:minmax_ineq}
\end{aligned}   
\end{equation}

Consequently, the right-hand side of inequality (\ref{eqn:minmax_ineq}) provides an upper bound for the primal optimization problem:

\begin{equation}
\min_\Theta \mathcal{L}(\Theta) \leq \max_\lambda \min_\Theta \mathcal{L}_d(\Theta, \lambda) = \bar{\mathcal{L}}.
\end{equation}

Given that the regularization term \(\|(\mathbf{I} - M)\Theta_\Delta\|^2\) is non-negative, the dual formulation does not exceed the value of the original loss function \(\mathcal{L}(\Theta)\), thereby establishing the desired upper bound. This ensures that the fine-tuning adheres to the imposed sparsity constraint, selectively updating parameters as dictated by the mask \(M\), while maintaining the stability and integrity of the LLM throughout the adaptation process.

\section{Details of the training mask ratio}
\label{sec:apdx_mask_ratio}

Table~\ref{tab:apdx_mask_ratio} describes the mask ratio details of the proposed \method{} method for all models and all domain experiments. To be fair in comparison,  One-off Drop, Random Mask and \method{} utilize the same mask ratio. All experiments reach optimal performance at 10\%-40\% sparsity updates.

\begin{table}[htbp]
\resizebox{\linewidth}{!}{%
\begin{tabular}{ccc}
\toprule
Model   & \multicolumn{1}{c}{Domain} & Mask Ratio (\%) \\
\midrule
\textsc{Mistral}-7B  & Math Reasoning &   20        \\
\textsc{Llama3}-8B  & Math Reasoning   &    30        \\
\textsc{Mistral}-7B & Code Generation      & 10           \\
\textsc{DeepSeek-Coder-Base-6.7B} & Code Generation &  30   \\          
\textsc{Llama2}-7B    & General     &    40        \\
\textsc{Llama2}-13B   & General     &  10 \\    
\bottomrule    
\end{tabular}%
}
\caption{Details of mask ratio for all experiments in this paper.}
\label{tab:apdx_mask_ratio}
\end{table}

\begin{figure}[t]
    \centering
    \includegraphics[width=1\linewidth]{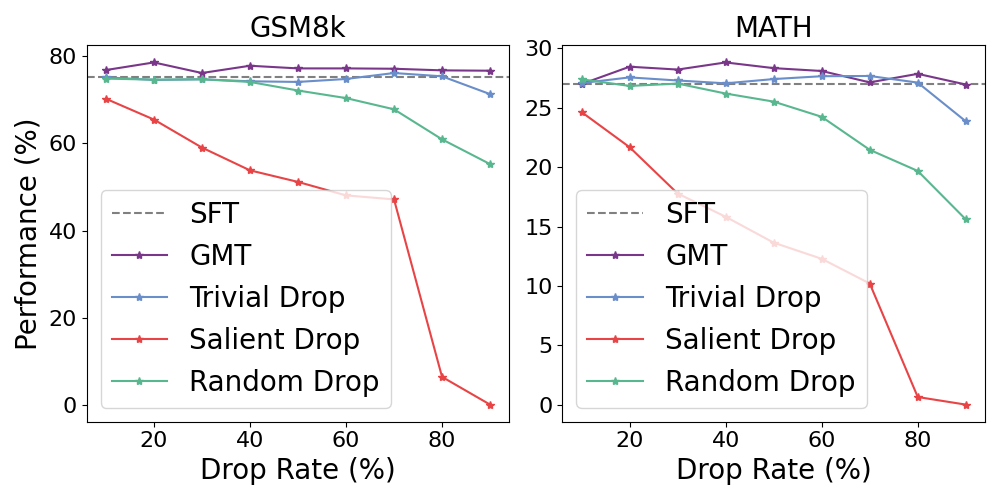}
    \caption{ We employed three strategies for selectively dropping delta parameters at various rates depending on the magnitude (absolute value) of the delta parameters: 1) preferentially dropping salient parameters, 2) preferentially dropping trivial parameters, and 3) dropping parameters randomly. }
    \label{fig:delta_drop_linechart}
\end{figure}

\end{document}